\documentclass[11pt]{article}

\usepackage[final]{acl}

\usepackage{times}
\usepackage{latexsym}
\usepackage[T1]{fontenc}
\usepackage[utf8]{inputenc}
\usepackage{microtype}
\usepackage{inconsolata}
\usepackage{graphicx}
\usepackage{breqn}

\usepackage{amsmath}
\usepackage{amsfonts}
\usepackage{booktabs}
\usepackage{tabularx}
\usepackage{multirow}
\usepackage{algorithm}
\usepackage{algorithmic}
\usepackage{float}
\usepackage[capitalize]{cleveref}   % load after hyperref (pulled in by acl)
\usepackage{orcidlink}

\newcommand{\ub}{\_\allowbreak}
\title{Business Compromise Detection with Agentic AI and LLM-driven Knowledge Discovery}

\author{
  \textbf{Diego Palma}~\orcidlink{0000-0003-1540-7164},
  \textbf{Kyu Bin Kim}~\orcidlink{0009-0005-2149-9642},
  \textbf{Zhen Han}~\orcidlink{0000-0002-6058-9203},
  \textbf{Allbright Dsouza}~\orcidlink{0009-0007-7895-0558},
\textbf{Zhiyuan Liu}~\orcidlink{0000-0001-8531-185X} \\
  Meta
\\
  \small{
    \textbf{Correspondence:} \href{mailto:dpalmasan@meta.com}{dpalmasan@meta.com}, \href{mailto:dpalmasan@meta.com}{zhiy@meta.com}
  }
}

\begin{document}
\maketitle

\begin{abstract}
Detecting compromised business ad accounts is a challenge in digital advertising, as attackers exploit hijacked accounts to launch fraudulent campaigns. Large Language Model (LLM) agents show promise for integrity enforcement, but hallucinated mistakes on hard cases create business friction. In a study we find the autonomous agent is a strong, recall-heavy \emph{signal extractor} but an unreliable \emph{final arbiter}, conceding precision on ambiguous decisions. We therefore keep the agent as an investigator that emits a structured, interpretable signal vector, and delegate the verdict to a neuro-symbolic stage: symbolic rules discovered by Inductive Logic Programming (FOIL-IE), a Na\"ive Bayes calibration layer, and a data-tuned contradiction layer. Evaluating on a compromise-over-sampled population and a realistic low-prevalence sample with subject-matter-expert labels, this \emph{arbiter substitution} raises MCC from 0.295 to 0.435 ($\Delta$MCC $+0.139$, 95\% CI $[+0.026,+0.245]$, $p=0.018$, paired bootstrap), lifting precision from 0.250 to 0.446 ($1.8\times$) at a recall cost (0.920 to 0.660). Benchmarked under identical conditions, it also edges human reviewers (0.422) and tree ensembles (0.386). The rules encode domain priors and generalize from few labels while remaining interpretable and auditable.

\end{abstract}

\section{Introduction}
\label{sec_intro}
The digital advertising ecosystem is increasingly targeted by malicious actors who exploit established ad accounts to launch fraudulent campaigns \cite{zhang2025ai,grier2012manufacturing}. When a legitimate business ad account is compromised, attackers bypass initial risk checks \cite{stone2011understanding} and leverage the account's historical reputation to run policy-violating ads \cite{lever2016domain}, causing financial loss and brand damage. Detection is hard because attackers constantly evolve their tactics, including synthetic profiles, IP obfuscation, and sock-puppet agency structures \cite{sakib2022automated}.

Traditional detection falls into two camps: rule-based systems \cite{zhang2008detecting,metwally2005using}, which are interpretable and actionable but brittle and need constant SME maintenance; and statistics-based models \cite{chicco2020advantages,zhang2025ai}, which generalize well but lack the interpretability needed to justify enforcement. More recently, LLM agents can autonomously reason and act \cite{he2024emerged}, and show promise for complex integrity violations. LLMs have the risk to hallucinate, which will be prohibitive for integrity violations as a wrong decision might disrupt customer operations and pose a risk that has to be minimized; while self-reflection \cite{yao2022react,shinn2023reflexion} reduces this risk, agents can still struggle to weigh subtle, contradictory business evidence and might behave inconsistently across runs. As a result, agents show a large precision gap in production, where the boundary between a compromised and a benign entity is genuinely ambiguous.

We propose a neuro-symbolic architecture that separates evidence extraction from the final decision. Following an expert mental model, the agent acts as an initial reviewer that (1) makes a decision and (2) reports the supporting evidence, which we convert into a structured signal vector. A neuro-symbolic stage then makes the decision in three layers: (L1) Inductive Logic Programming:  a combination of the First-Order Inductive Learner (FOIL) and Inverse Entailment (IE), that automatically discovers interpretable rules; (L2) a probabilistic layer that models the conditional probabilities of those rules to handle borderline cases; and (L3) a contradiction layer that reconciles conflicting evidence to rescue false negatives and veto false positives.

Our contributions are:
\begin{enumerate}\itemsep0pt
\item An empirical finding: an autonomous LLM agent is an effective recall-heavy \emph{signal extractor} but a poor \emph{final arbiter} on ambiguous cases, motivating a separation of extraction (LLM) from decision (a knowledge stage) that we argue generalizes to other integrity problems.
\item An LLM agent that emits both a decision and a structured signal vector aligned with an expert mental model, turning a stochastic free-text reviewer into a deterministic, interpretable feature source.
\item A three-layer neuro-symbolic decision stage that discovers interpretable rules (FOIL-IE), calibrates them in a probabilistic fashion, and arbitrates contradictions. Benchmarked under \emph{identical conditions} against tree ensembles, it attains the highest point estimates on realistic low-prevalence data; the improvement over the agent is statistically significant, while the margins over human reviewers and the ensembles are not (\cref{sec_exp_holdout}). A single interpretable tree is markedly weaker throughout.
\item An evaluation on two populations (an over-sampled general population and a realistic, expert-labeled hard-case sample), with paired-bootstrap confidence intervals, a per-method error analysis, and a threshold sensitivity study that together identify which of our design choices the data actually supports, including the contradiction \emph{resolution} policy, that it does not (\cref{sec_l3_decomp}).
\end{enumerate}

\section{Background}
\label{sec_related_work}
Business compromise is among the hardest integrity problems: compared to generic fraud, the boundary between violating and benign is ambiguous, and reviewers given the same contradictory evidence often disagree (e.g., distinguishing a legitimate agency from a compromised account). Fraud detection has historically driven automation under severe class imbalance regime \cite{zojaji2016survey}, and graph-based methods such as inductive graph representation learning improve performance by embedding transaction/entity networks \cite{van2022inductive}. Graph signals like connection ages and shared business entities are useful, but these models rarely expose them in an interpretable form.

To gain interpretability, Inductive Logic Programming (ILP) \cite{muggleton1991inductive} and Inverse Entailment \cite{cropper2022inductive}, together with Quinlan's relational learning \cite{quinlan1990learning}, learn human-readable rules. Recent work combines logic with statistics: learning rules from noisy data \cite{evans2018learning}, Probabilistic Soft Logic \cite{bach2017hinge}, and Differentiable ILP for fraud \cite{wolfson2024differentiable}, which finds that explainability comes at a scalability cost relative to decision trees \cite{wu2008top} and that logic generation paired with probabilistic scoring can close the gap. We adopt exactly this combination: a fast FOIL-IE layer to generate discrete rules and a Na\"ive Bayes layer for probabilistic robustness. Separately, LLM agents \cite{yao2022react,shinn2023reflexion,wei2022chain,kojima2022large} have advanced autonomous reasoning and tool use, but building a reliable \emph{decision} system from them remains hard in domains, like business compromise, that require imposing domain knowledge: the gap our neuro-symbolic stage targets.

% TODO(camera-ready): regenerate as vector art with larger type (reviewer 43Fo).
% Source artwork is not in this repo. Requirements: true vector PDF (no raster
% embed), body text >= 8pt at final column width, and legible at 100% zoom in
% greyscale print.
\begin{figure*}[t]
  \centering
  \includegraphics[width=\textwidth]{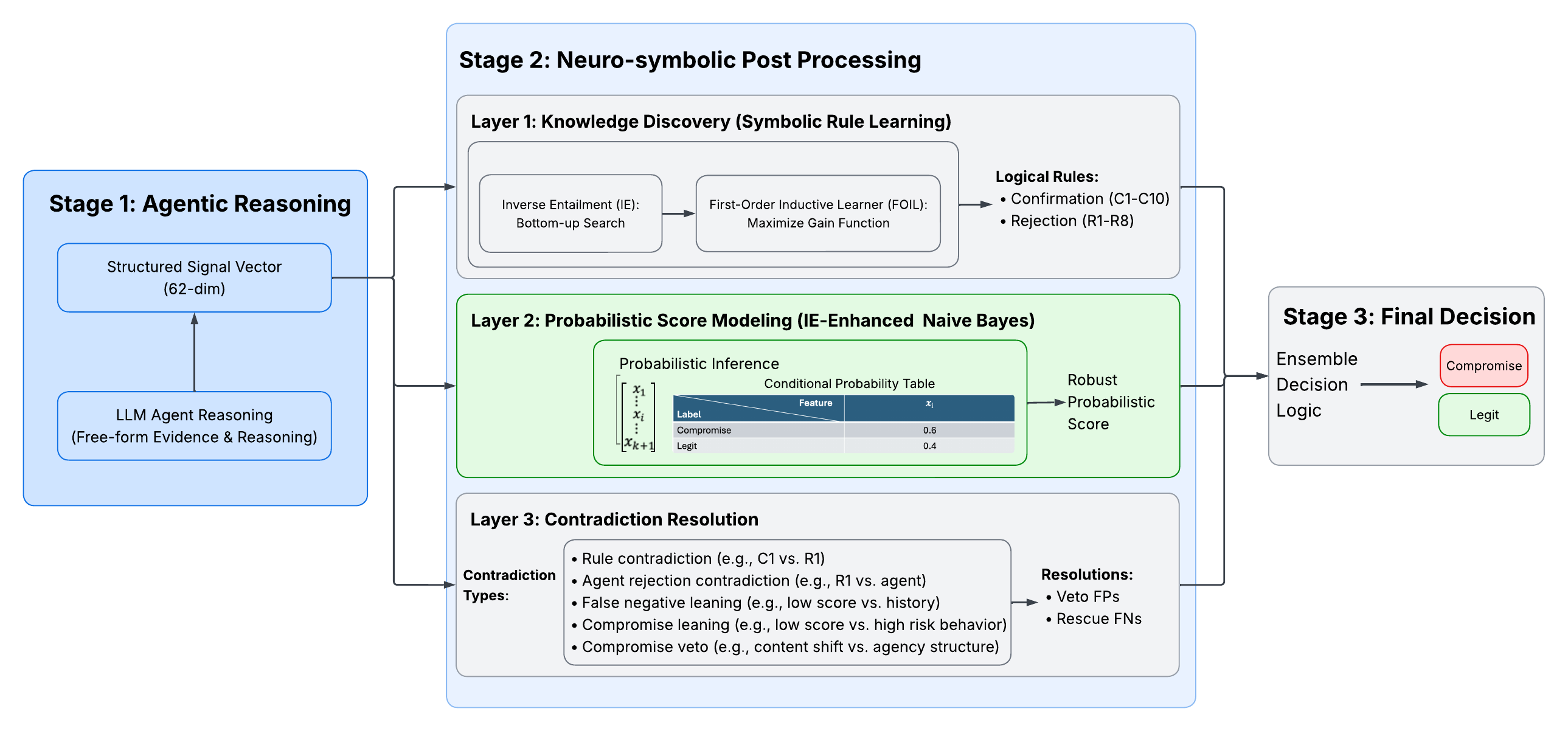}
  \caption{Three-stage approach. Stage 1: an LLM agent inspects business signals and yields a decision plus reasons, which we convert into a structured signal vector. Stage 2: a neuro-symbolic decision stage of three layers (FOIL-IE rules, Na\"ive Bayes, contradiction resolution). Stage 3: the final decision.}
  \label{fig_method_overview}
\end{figure*}

\section{Methodology}
\label{sec_approach}
We focus on the two modules (\cref{fig_method_overview}): a multi-turn LLM agent over a rich toolset (\cref{sec_llm_agent}) and a neuro-symbolic decision stage (\crefrange{sec_foil}{sec_contradiction}). \Cref{app_walkthrough} traces one real account end to end: agent trace, JSON verdicts, signal vector, L1, L2, L3, decision. For readers who prefer to see the pipeline concretely before the component descriptions.

\subsection{Terminology}
\label{sec_terminology}
Four account states are easily conflated, and the distinction drives both our labels and our dominant error mode. A \textbf{legitimate} account is operated within policy by its rightful owner. An \textbf{abusive} account is one where the \emph{rightful owner} runs policy-violating campaigns, so control has not changed hands; a \textbf{compromised} account is a legitimate one \emph{taken over by a third party}, who then runs campaigns the owner did not authorize. \textbf{Fraudulent} describes any policy-violating campaign regardless of who controls the account, so both abusive and compromised accounts produce fraudulent ads. Our target is compromise specifically, because it triggers a different enforcement path (account recovery and owner notification) than abuse (owner-directed penalties). The two are hard to separate from the ad alone, both yield violating content, and confusing \emph{abusive} for \emph{compromised} is the single most frequent error across every method we evaluate (\cref{tab_error_analysis}).

\subsection{LLM Agent for Business Compromise}
\label{sec_llm_agent}
We build the agent around an \emph{expert mental model}: a structured checklist, elicited from subject matter experts (SMEs), of the questions a reviewer asks when adjudicating an account (e.g., "is the destination domain newly registered?", "does the ad creator behavior match the account creators history?"), together with the policy conditions under which an answer argues for or against compromise. We phrase these questions to have objective, evidence-grounded answers, so the agent's job is to \emph{gather and report evidence} rather than exercise open-ended judgment.

Each question is backed by a tool, exposed via the Model Context Protocol (MCP), an open standard that lets an LLM call external functions and consume their structured results within a reasoning loop. Our tools span different evidence families (e.g. destination URL/domain age, targeting behaviors, creator profile and behavior, page history, business-relationship structure, payment patterns, account-connection timing, among others). Tools are granular and single-purpose so the agent fetches only what it needs and revises its reasoning as evidence arrives. We also prompt the agent with tool priorities so it can resolve conflicts (e.g. a behavior-shift signal vs.\ a verified agency structure) following the expert mental model.

\paragraph{Agent configuration.} \Cref{tab_agent_config} in \cref{app_prompt} gives the full specification. The backbone is Gemini 2.5 Flash on a hosted serving platform, driven through native function calling; tools are discovered at runtime over MCP. The loop is multi-turn and evidence-revising: each turn the model may request one or more tool calls, the structured results are appended to the dialog, and the loop recurses until the model returns a final answer with no tool call, or until a hard cap of 25 steps. Identical tool calls are memoized within a run, so an account yields the same signal vector across runs. The system prompt is the expert mental model made explicit: an expert persona and threat context, the compromise-vs-abuse-vs-legitimate distinction of \cref{sec_terminology} with a "default to Not Compromise" prior, a 9-component review checklist, and 8 mandatory override checks. Chain-of-thought is required in a free-text field, and the final answer is a single JSON object. The same appendix gives the prompt's block structure and the full output schema; the prompt's policy content (the adversary monetization taxonomy and the conditions and thresholds of the override checks) is proprietary enforcement material and is withheld, which we record as a reproducibility limitation.

The agent's structured output is a set of small categorical verdicts (per-family judgments in $\{$benign, inconclusive, suspicious$\}$) consumed as logic predicates, joined by boolean evidence signals (e.g.\ \emph{domain newly registered}) and bounded numerics (a derived risk score, account connection ages). Committing the agent to this vocabulary makes hallucinations surface as \emph{inconsistency} between predicates rather than as fluent prose, and lets the downstream stage re-weight checks the agent over-trusts. Because every dimension is grounded in a deterministic tool result, the same account yields the same vector across runs, stabilizing downstream learning. This matters because easy cases, where one or two checks settle the decision, are already high-precision for the agent alone; it is the mid-band \emph{hard cases} that produce the four recurring failure modes of \cref{tab_misclassify}: abusive advertisers read as compromised, multi-vertical agencies read as compromised, benign behavioral shifts read as compromised, and outright hallucinations. Most stem from missing domain knowledge, motivating the knowledge stage below.

\subsection{Layer 1: Symbolic Learning with FOIL-IE}
\label{sec_foil}
We use ILP \cite{muggleton1991inductive} to learn symbolic rules that separate positive from negative examples, choosing FOIL \cite{quinlan1990learning}. To overcome FOIL's connectivity limitation \cite{quinlan1991determinate}, we integrate Inverse Entailment \cite{muggleton2001completing}, forming FOIL-IE. Learning maximizes the gain

\begin{dmath}\label{eq_gain_func}
    Gain(l, h) = t \cdot \left( \log_2 \frac{p(l)}{p(l) + n(l)} - \log_2 \frac{p(h)}{p(h) + n(h)} \right)
\end{dmath}

where $h$ is the current rule, $l$ a candidate literal, $p(\cdot)$ and $n(\cdot)$ the positive/negative coverage, and $t$ the positives covered by $h$ that remain covered after adding $l$. With background knowledge $B$ and examples $E^+,E^-$, the procedure (\cref{alg_foil_ie}, appendix) yields a rule set $H$: confirmation rules that recognize compromise, and rejection rules that recognize legitimate accounts.

\subsection{Layer 2: Probability Modeling}
\label{sec_nb}
To handle borderline cases, we associate each rule $h_i\in H$ with class-conditional probabilities via a probabilistic layer. For the case of this paper, we used Na\"ive Bayes due to the compromise prevalence being low in this problem space. Each rule becomes a boolean feature (a first-order signal); we also include raw signals (zeroth order), rule combinations (second order), and contradiction flags (third order). Conditional probabilities are stored in a Conditional Probability Table \cite{Koller2009} with Laplace smoothing \cite{Manning2008}:
\begin{equation}P(x_i \mid y) = \frac{N_{yi} + \alpha}{N_y + 2\alpha}\end{equation}
where $N_{yi}$ counts feature $x_i$ in class $y$, $N_y$ is the class size, and $\alpha{=}1$. Feeding the L1 rule-fires back as features (the IE enhancement) lets L2 calibrate the discrete rules probabilistically.

\subsection{Layer 3: Contradiction Resolution}
\label{sec_contradiction}
Hard cases often present contradictory evidence, made worse by agents hallucination. We are deliberate about what is learned here, and state the division plainly.

\noindent\textbf{The categories are an expert prior, not a discovery.} We define five contradiction categories: compromise-leaning, false-negative-leaning, false-positive-leaning, direct rule contradictions, and agent-rejection contradictions (\cref{tab:contradiction-types}, appendix). These are not learned. They encode an expert prior over the four failure modes we observed in (\cref{tab_misclassify}), and we claim no more for them than that.

\noindent\textbf{The inputs are not hand-set.} Direct and agent-rejection contradictions are computed from the FOIL-IE rules discovered in L1; the leaning flags come from the deterministic MCP signals plus the L2 posterior.

\noindent\textbf{The resolution policy is two swept thresholds.} A false-positive-leaning flag vetoes a positive decision when the posterior is low, and a false-negative- or compromise-leaning flag rescues a negative decision when the rules do not fire and the posterior sits just below threshold. The veto and rescue thresholds are swept on training folds for maximum MCC, the same protocol that selects tree depth and ensemble size for our baselines, never on test data.

Being a second-order layer, L3 is the most data-hungry component. In \cref{sec_l3_decomp} we show that at our current label volume the resolution policy is in fact \emph{static}: the gain we previously attributed to L3 comes from feeding the contradiction flags to L2 as features and from the posterior floor, not from the veto/rescue actions.

\section{Experiments and Results}
\label{sec_exp}
We evaluate each stage and layer (\cref{sec_approach}): setup (\cref{sec_exp_setup}), cross-validation on a class-rich population (easy and hard cases) (\cref{sec_exp_cross_validate}), and an ablation against humans and tree ensembles on hard cases (lower compromise prevalence) (\cref{sec_exp_holdout}).

\subsection{Experimental Setup}
\label{sec_exp_setup}
\noindent\textbf{Data.} We use two populations. The first is an \emph{over-sampled general population}: 975 labeled cases drawn from the broad history of past reviews rather than from the hard cases we ultimately target, with compromised accounts over-sampled ($\sim$70\% positive) to obtain a class-rich set for higher-powered model comparison and rule learning, evaluated with 5-fold stratified cross-validation. We rely on this over-sampled set precisely because high-confidence labels for the \emph{hard} cases scarce; it is not prevalence-matched to production and we treat it as a controlled, higher-power comparison rather than an estimate. The second is the \emph{production hard-case population}: 397 accounts routed to the agent and adjudicated by SMEs, of which 60 ($15.1\%$) are compromised; the operationally faithful, low-prevalence target. We also consider an independent human review (non-SME, trained and outsourced), and the comparison is the system that was trained on SME labels vs these non-SME  reviewers. \Cref{tab:ablation-train} and every significance test are computed on this subset. A defining constraint is \emph{label scarcity}: a high-confidence compromise label requires an SME to reconcile contradictory evidence, and even trained human reviewers agree with SME adjudication only 75-85\% of the time (\cref{sec_exp_holdout}). Such labels are slow and costly, bounding dataset size and favoring sample-efficient, interpretable methods.

\noindent\textbf{Comparative methods.} (1) an always-positive baseline, which predicts compromise for every case (F1 is high only because the over-sampled population is 70\% positive; its MCC is 0 by construction, which is the point of including it); (2) the LLM agent alone; (3) FOIL-IE only (L1); (4) Na\"ive Bayes only (L2); (5) ILP+NB (L1+L2); (6) the full neuro-symbolic stage (L1+L2+L3); and three \emph{tree baselines} on the same structured signals: (7) a single decision tree (CART), (8) random forest, (9) gradient-boosted trees. To ensure fairness, every learned method uses the same folds, the same agent gating, and per-fold decision-threshold tuning; tree depth/ensemble size are selected by the same cross-validated MCC we use to tune our own thresholds.

\noindent\textbf{Metrics.} We report recall (capturing compromise early limits loss), F1, and the Matthews Correlation Coefficient (MCC) \cite{chicco2020advantages}, which is robust under imbalance ($0$ = random).

\subsection{Over-sampled General Population}
\label{sec_exp_cross_validate}
\Cref{tab:cv-simulated} in \cref{app_cv} reports 5-fold CV. First, \emph{every} knowledge layer dramatically improves precision over the agent (0.809 $\rightarrow$ 0.92-0.95) and lifts MCC from 0.513 to 0.74-0.82: the agent is a strong recall-heavy extractor, and the value comes from re-arbitrating its decision. Second, on this class-rich population \emph{tree ensembles are competitive}: random forest attains the best MCC (0.823), marginally above our stage (0.807), with gradient boosting tied (0.800), gaps within CV and tuning noise. We report this honestly because it does \emph{not} hold on realistic data (\cref{sec_exp_holdout}): the ranking reverses, since abundant positives are exactly where high-capacity ensembles thrive. Even here, the only interpretable tree (a single decision tree) is markedly weaker (0.733), so an ensemble buys accuracy at the cost of auditable rules.

\begin{table*}[t]
  \centering
  \caption{Comparison of different approaches to ground truth (SME).}
  \label{tab:ablation-train}
\small
  \begin{tabular}{llcccccc}
  \toprule
  \textbf{Model} & \textbf{Layers} & \textbf{Prec} & \textbf{Rec} & \textbf{F1} & \textbf{MCC} & \textbf{MCC 95\% CI} & \textbf{MCC per-fold SD} \\
  \midrule
  Human Reviewer & N/A & 0.510 & 0.520 & 0.515 & 0.422 & [0.286, 0.553] & --- \\
  Agent Only & N/A & 0.250 & \textbf{0.920} & 0.393 & 0.295 & [0.214, 0.369] & 0.089 \\
  Gradient Boosting & N/A & \textbf{0.487} & 0.380 & 0.427 & 0.338 & [0.192, 0.479] & 0.135 \\
  Random Forest & N/A & 0.450 & 0.540 & 0.491 & 0.386 & [0.252, 0.513] & 0.065 \\
  \midrule
  FOIL-IE & L1 & 0.425 & 0.680 & 0.523 & 0.425 & [0.304, 0.540] & 0.154 \\
  NB Only & L2 & 0.426 & 0.520 & 0.468 & 0.358 & [0.227, 0.486] & 0.145 \\
  ILP+NB & L1$+$L2 & 0.375 & 0.720 & 0.493 & 0.391 & [0.277, 0.501] & 0.130 \\
  Full & L1$+$L2$+$L3 & 0.446 & 0.660 & \textbf{0.532} & \textbf{0.435} & [0.309, 0.553] & 0.147 \\
  \bottomrule
  \end{tabular}
\end{table*}

\begin{table}[t]
  \centering
  \caption{Paired bootstrap ($B{=}10{,}000$, shared resample indices) of the full stage minus each baseline}
  \label{tab_significance}
\small
\setlength{\tabcolsep}{4pt}
  \begin{tabular}{@{}lcc@{}}
  \toprule
  \textbf{Full stage vs.} & \textbf{$\Delta$MCC [95\% CI]} & \textbf{$p$} \\
  \midrule
  LLM Agent alone & $+0.139$ $[+0.026, +0.245]$ & \textbf{0.018} \\
  Gradient Boosting & $+0.097$ $[-0.041, +0.236]$ & 0.163 \\
  L2: NB only & $+0.076$ $[-0.037, +0.190]$ & 0.187 \\
  Random Forest & $+0.049$ $[-0.073, +0.171]$ & 0.437 \\
  L1$+$L2: ILP$+$NB & $+0.043$ $[-0.037, +0.120]$ & 0.275 \\
  Human Reviewer & $+0.012$ $[-0.155, +0.174]$ & 0.881 \\
  L1: FOIL-IE & $+0.010$ $[-0.056, +0.071]$ & 0.768 \\
  \bottomrule
  \end{tabular}
\end{table}

\subsection{Low Prevalence Population}
\label{sec_exp_holdout}
We now turn to the operationally faithful population at the true $\sim$15\% prevalence: a deliberately hard sample where human reviewers agree with SME ground truth only 84\% of the time (75\% on an earlier curated subset), attaining MCC 0.422. To compare every method against humans \emph{on equal footing}, \cref{tab:ablation-train} restricts to the 397 accounts with both an SME label and a human review, running every learned method through identical folds, agent gating, and per-fold threshold tuning, so differences reflect the model, not the protocol.

\noindent\textbf{The supported result is the arbiter substitution.} Holding the agent's evidence fixed and replacing only \emph{who decides}, the agent's own verdict versus the knowledge stage, raises MCC from 0.295 to 0.435 and precision from 0.250 to 0.446, a factor of $1.8$, moving the recall from 0.920 to 0.660. In absolute terms the stage removes 97 of the agent's 138 false positives on these 397 cases while giving up 13 true positives. This is the one comparison in the paper that survives a paired bootstrap ($\Delta$MCC $+0.139$, 95\% CI $[+0.026, +0.245]$, $p=0.018$; \cref{tab_significance}), and it is the claim the paper rests on.

\noindent\textbf{The stage achieves slightly higher MCC than the evaluated human reviewers, but the difference is not significant.} The full stage reaches MCC 0.435 against the human reviewer's 0.422 ($\Delta$MCC $+0.012$, $p=0.881$). We therefore do not claim to exceed human experts. The human number is better read as context for how hard this slice is: trained reviewers match SME adjudication only 84\% of the time, so a system that merely \emph{matches} them while being interpretable and fully automatic is already operationally useful.

\noindent\textbf{Ahead of tree ensembles in point estimate, within noise in significance.} Under identical conditions, random forest (0.386) and gradient boosting (0.338) fall below the knowledge-based methods, reversing their ranking on the over-sampled general population. The proposed mechanism is sample efficiency: with only 50 positives the ensembles overfit, whereas the FOIL-IE rules encode SME domain priors and generalize from few examples. We report this as the more likely explanation rather than a demonstrated one (the margin over random forest is not significant ($\Delta$MCC $+0.049$, $p=0.437$), and confirming sample efficiency properly needs a learning curve over label volume, which our label budget does not yet support.

\noindent\textbf{The ablation is monotone in point estimate only.} Na\"ive Bayes alone (L2) scores 0.358; adding the discovered rules (L1+L2) raises it to 0.391; the full stage reaches 0.435. None of these steps is individually significant in this dataset (\cref{tab_significance}), see \cref{sec_l3_decomp}.

\subsection{What L3 Actually Contributes}
\label{sec_l3_decomp}

The full stage differs from L1+L2 in \emph{two} ways at once, which the original ablation conflated: it swaps the raw Na\"ive Bayes posterior for the IE-enhanced one (rule fires and contradiction flags as \emph{features}), \emph{and} it applies the posterior floor and the veto/rescue \emph{actions}. \Cref{tab_l3_decomp} separates them. The IE-NB feature enhancement contributes $+0.025$ MCC and the posterior floor a further $+0.019$. The veto and rescue actions contribute \emph{exactly nothing}: enabling either or both leaves all four metrics bit-identical.

This is corroborated by the tuner and by a sensitivity sweep. Per-fold threshold selection chooses rescue-off in 5 of 5 folds and veto-off in 3 of 5. Sweeping the veto threshold over $\{0.0, 0.4, 0.5, 0.6\}$ and the rescue threshold over $\{$off$, 0.4, 0.5, 0.6\}$ (16 configurations, thresholds pinned rather than tuned) yields only two distinct outcomes: MCC 0.435 for veto $\leq 0.4$ and 0.420 for veto $\geq 0.5$, with the rescue threshold making no difference anywhere. The policy is flat, not knife-edge, which is reassuring for stability.

The honest reading is that at 50 positives the contradiction categories earn their place as \emph{features} for the probabilistic layer, and the posterior floor earns its place as a precision guard, but the second-order \emph{resolution policy} is static. L3 is the most data-hungry component of the design and we do not yet have the labels to justify it. We keep it in the architecture because it is the component we expect to activate first as SME labels accrue, and we flag it as unjustified at present rather than quietly reporting the aggregate.

\begin{table}[t]
\centering
\caption{Decomposing the full stage on the hard cases.}
\label{tab_l3_decomp}
\small
\setlength{\tabcolsep}{4pt}
\begin{tabular}{@{}lcccc@{}}
\toprule
\textbf{Configuration} & \textbf{Prec} & \textbf{Rec} & \textbf{F1} & \textbf{MCC} \\
\midrule
L1+L2: ILP $\vee$ raw-NB & 0.375 & 0.720 & 0.493 & 0.391 \\
A: ILP $\vee$ IE-NB & 0.407 & 0.700 & 0.515 & 0.416 \\
B: A $+$ posterior floor & 0.446 & 0.660 & 0.532 & 0.435 \\
C: B $+$ veto & 0.446 & 0.660 & 0.532 & 0.435 \\
D: B $+$ rescue & 0.446 & 0.660 & 0.532 & 0.435 \\
E: B $+$ veto $+$ rescue & 0.446 & 0.660 & 0.532 & 0.435 \\
\bottomrule
\end{tabular}
\end{table}

\subsection{Per-Method Error Analysis}
\label{sec_error_analysis}
\Cref{tab_error_analysis} breaks each method's errors into the four failure modes of \cref{tab_misclassify}. False positives are bucketed from the agent's own signal vector (first match wins: \emph{abusive} if the abuse override fired, else \emph{agency} if the account is an agency, else \emph{benign shift} if a content, language, or target-country shift was detected, else \emph{other}), so the breakdown is reproducible from released signals rather than a fresh manual re-read.

Three things stand out. First, \textbf{benign shifts dominate every method's false positives}, 109 of the agent's 138, and still 32 of the full stage's 41. A legitimate business changing its creative, language, or target market is the hardest negative in this problem, and no layer solves it; the knowledge stage merely removes two thirds of these cases. Second, \textbf{the abusive-vs-compromised confusion is the error the knowledge stage fixes best}: the agent makes 10, the full stage 4. This is the confusion \cref{sec_terminology} defines and the one with the most costly enforcement consequence, so it is the right one to have improved. Third, \textbf{tree ensembles buy their precision by giving up recall}: gradient boosting has the fewest false positives of any method (20) but misses 31 of 50 compromises, versus 17 for the full stage. Under a policy where a missed compromise keeps an attacker live on a real business's account, that trade is not obviously the right one, and MCC alone obscures it.

\begin{table}[t]
\centering
\caption{Errors split by category.}
\label{tab_error_analysis}
\small
\setlength{\tabcolsep}{3pt}
\begin{tabular}{@{}lcccc@{\hskip 6pt}c@{}}
\toprule
& \multicolumn{4}{c}{\textbf{False positives}} & \\
\cmidrule(r){2-5}
\textbf{Method} & \textbf{Abus.} & \textbf{Agcy.} & \textbf{Shift} & \textbf{Other} & \textbf{FN} \\
\midrule
Agent alone & 10 & 10 & 109 & 9 & 4 \\
L1: FOIL-IE & 4 & 3 & 35 & 4 & 16 \\
L2: NB only & 5 & 4 & 25 & 1 & 24 \\
L1+L2: ILP+NB & 6 & 6 & 44 & 4 & 14 \\
Full (ours) & 4 & 4 & 32 & 1 & 17 \\
Random Forest & 0 & 3 & 29 & 1 & 23 \\
Gradient Boosting & 0 & 2 & 17 & 1 & 31 \\
\bottomrule
\end{tabular}
\end{table}

\subsection{Operational Characteristics}
\label{sec_cost}
The decision stage is operationally negligible: rule evaluation, the Na\"ive Bayes posterior, and arbitration together take 21\,ms on CPU (p99 25\,ms), four orders of magnitude below the agent, and re-fitting the entire stage on fresh labels is minutes of single-machine compute. Most of the execution time is essentially all agent-side, bounded above by the 25-step cap. Over 330 reviews, a review takes 175\,s on average while emitting only $\sim$740 estimated tokens, so wall-clock is dominated by sequential tool round-trips rather than by generation, and would shrink with tool parallelism rather than a faster decoder. Tool use is heavily skewed (median 1 call, mean 3.6, p99 16): the checklist is not run exhaustively, most accounts are resolved by the first pieces of evidence, and the step cap acts as a tail guard rather than the operating point. \Cref{app_cost} reports the full distributions.

\noindent\textbf{Where the human stays in the loop.} In execution the stage decides automatically, with the false-positive guard of \cref{sec_l3_decomp} applied before any enforcement action. A human reviewer is retained only for a subset of managed accounts, where a false positive is most damaging to a real business. Humans are therefore required for labeling and evaluation, SME adjudication is our ground truth, and appears as a baseline throughout, but not for routine decisions.

\section{Conclusion}
\label{sec_conclude}
We presented a system that pairs an LLM agent with a neuro-symbolic decision stage for business compromise. The agent is a strong evidence extractor but a poor arbiter, and the supported finding is the substitution itself: keeping the agent's evidence and replacing only the decision-maker raises MCC from 0.295 to 0.435 on realistic low-prevalence data ($p=0.018$), lifting precision $1.8\times$ at a recall cost. The stage achieves slightly higher MCC than the evaluated human reviewers and than tree ensembles, but neither margin is significant at this sample size and we do not claim them; on class-rich data ensembles are ahead. Decomposing our own stage shows the contradiction categories earn their place as probabilistic features while the veto/rescue policy they were designed for is currently static, it is a component the architecture is ready for but the label volume does not yet justify. The discovered rules align with SME reasoning and remain auditable. We next aim to instrument raw trace capture so the agent baselines discussed in Limitations become runnable, distill the agent into a smaller model for throughput, and test the paradigm on other integrity problems.

\section*{Limitations}
We are candid about several limitations. First, on the over-sampled general population, tree ensembles match or slightly exceed our stage on raw accuracy (random forest MCC 0.823 vs.\ 0.807); there, our advantage is interpretability rather than accuracy. On the realistic population our point estimates are highest, but only the margin over the agent is statistically significant (\cref{tab_significance}), so the ranking against humans and ensembles should be read as unresolved rather than in our favor. Second, the contradiction layer (L3) is the most data-hungry component, and at the 50 positives available in the matched subset its veto/rescue policy contributes nothing measurable (\cref{sec_l3_decomp}); we retain it as architecture, not as a validated result. Sample efficiency, which we propose as the mechanism behind the knowledge stage's advantage on low-prevalence data, is likewise a proposed rather than a demonstrated explanation, absent a learning curve over label volume. Third, our sample is small and expensive to grow, SME-adjudicated accounts (hard cases subset). Absolute numbers therefore carry wide confidence intervals and should be read as relative comparisons under matched conditions. Fourth, all results are from a single platform and a single integrity problem; while we argue the paradigm generalizes, we have not measured it on another domain.

\paragraph{Stronger agent baselines are the main open gap.} We compare against the agent in one setting, not against self-consistency, reflection, verifier-based reasoning, or majority voting. Two of these are partially addressed by design (the agent already runs a multi-turn, evidence-revising loop with structured output, and our L2/L3 stage is a learned verifier over its structured evidence) but that is an argument, not a measurement. The obstacle to a controlled study is real: several tools return time-dependent results (account state, connection ages, recent activity), so a fair multi-run comparison requires replaying an identical frozen evidence snapshot to every run. Our traces retain the parsed signal vector but not the raw tool outputs, so even the frozen-trace self-consistency variant we offered in review is not currently runnable; recovering raw traces is an instrumentation change we have scoped but not shipped. \textbf{Consequently our claim that a knowledge stage beats better prompting is untested, not confirmed.} It is the experiment we would run first in as a follow up study.

\paragraph{Backbone robustness is untested.} The decision stage learns over the structured signal vector rather than raw LLM text, so decision quality should be largely decoupled from the extractor's identity, and the backbone is a pluggable component in our implementation. We have not verified this with a controlled two-backbone comparison, for the same frozen-evidence reason.

\paragraph{The prompt is not fully reproducible.} \Cref{app_prompt} gives the block structure, the complete output schema, and the full agent configuration, which is enough to re-implement the architecture but not to reproduce our exact extractor. The components a third party can reproduce exactly are the decision stage and its evaluation protocol, given a signal vector of the same shape.

\paragraph{Adversarial robustness and updating.} Compromise is adversarial, so static models decay. Two design properties help: because the agent only extracts evidence and a cheap stage decides, adapting to a new tactic means re-fitting the symbolic stage on fresh labels (minutes) rather than re-prompting the agent; and the rules are human-readable, so SMEs can inspect, retire, or add them. Our results already show distribution sensitivity, the method ranking changes between populations (tree ensembles lead on class-rich data, the neuro-symbolic stage on realistic low-prevalence data), so a model tuned at one operating point should not be assumed optimal at another. During execution we monitor per-rule precision and firing rates, periodically re-learn rules and thresholds as labels accrue, and treat a sustained precision drop as a drift alarm. Quantifying decay rate and re-fit cadence needs a longer longitudinal label stream than we currently have.

\paragraph{Latency and operational characteristics.} We optimize for review quality, not serving. Measured per-review figures are in \cref{sec_cost}: the dominant characteristic is the agent's multi-turn tool calling, and, within it, sequential tool latency rather than token generation, while the neuro-symbolic stage adds 21\,ms. The scaling path we are pursuing is to distill the multi-turn agent into a smaller task-specialized model that emits the signal vector directly, leaving the interpretable decision stage unchanged.

\section*{Acknowledgments}
We are grateful for many conditions and people that make this work available. Our work relies on the platforms that provide necessary foundation for this work. With all these conditions, we are able to conduct, drive and deliver this work. 

We thank our colleagues and leaders who supported and contributed to this work. Specifically, we thank Nicola Bortignon, Cindy Liu, Bhagya Bethala, Xuefu Wang, Romano Adler, Guang Yang, Lewis Woodease, Arun Chandrasekanran, Mark Atherton for their insightful suggestions and support. We thank Lia Xu for the analysis in production.

\bibliography{ref}

\appendix

\section{FOIL-IE Algorithm}
\label{app_alg}
\Cref{alg_foil_ie} details the hybrid FOIL-IE procedure referenced in \cref{sec_foil}.

\begin{algorithm}[H]
\caption{Hybrid FOIL-IE Algorithm}
\label{alg_foil_ie}
\begin{algorithmic}[1]
\REQUIRE $B, E^+, E^-$
\ENSURE $H$ (the set of learned rules)
\STATE $H \gets \emptyset$;\quad $E_{rem} \gets E^+$
\WHILE{$E_{rem} \neq \emptyset$}
    \STATE \textbf{Selection:} pick an example $e \in E_{rem}$
    \STATE \textbf{Saturation (IE):} build bottom clause $\perp_e \gets \{ l \mid B \land \neg e \models \neg l \}$
    \STATE \textbf{Reduction (FOIL):} find $h \preceq \perp_e$ maximizing $Gain(l,h)$ (\cref{eq_gain_func})
    \STATE \textbf{Update:} $H \gets H \cup \{h\}$;\ \ $E_{rem} \gets E_{rem} \setminus \{ e' \mid B \cup \{h\} \models e' \}$
\ENDWHILE
\RETURN $H$
\end{algorithmic}
\end{algorithm}

\section{Agent Mis-classification Categories}
\label{app_misclassify}

\begin{table}[t]
\centering
\caption{Four recurring agent mis-classifications (all over-predict compromise).}
\label{tab_misclassify}
\small
\begin{tabularx}{\linewidth}{clX}
\toprule
\# & Expert & Mistake reason\\
\midrule
1 & Abusive & Confused abusive with compromised; over-weighted abuse signals.\\
2 & Benign & Over-weighted a content-shift signal.\\
3 & Benign & Confused a legitimate agency with a compromise.\\
4 & Benign & Hallucination: recognized agency signals yet decided compromise.\\
\bottomrule
\end{tabularx}
\end{table}

\section{Worked Example: One Account End to End}
\label{app_walkthrough}
This traces a single account through every stage, as promised in review. It is a real case from the "hard cases" set, de-identified: no account, business, domain, or country is named, and every value shown is drawn from the released signal vector. It was chosen because it is a genuine hard case where the human reviewer called it Not Compromise, the SME called it Compromise, and the full stage recovered it.

\noindent\textbf{Stage 1: agent.} The agent works the checklist, calling tools per component. Its structured output records \emph{suspicious} on four components (content, destination URL, page, target countries), \emph{normal} on creator and business connections, and \emph{inconclusive} on the remaining four, giving \texttt{num\ub suspicious\ub checks} $=5$ with \texttt{confidence} high.

\noindent\textbf{Signal vector.} The categorical verdicts and booleans above are joined by the deterministic tool numerics: risk score $0.39$ (squarely in the mid-band where the agent is unreliable), total risk $5.0$, a destination domain new to this business though 416 days old, 8 permitted users, all ad accounts under one business, no prior page use, no compromised pixel, and zero baseline ads.

\noindent\textbf{L1.} The confirmation rules fire on the conjunction of a mid-band score with a language/target-country shift on an account carrying no baseline ad history; no rejection rule fires, because the two override checks that would normally protect an established advertiser are contradicted by the zero baseline. \texttt{ilp\ub comp} is true.

\noindent\textbf{L2.} The IE-enhanced posterior sits above the tuned decision threshold: the rule-fire features and the shift-without-congruence combination both push upward, while \emph{creator is original to business} pushes down; this is not a creator-takeover signature but a session-takeover one.

\noindent\textbf{L3.} No contradiction category fires: there is no rule conflict, the agent and the rules agree, and no false-positive-leaning evidence is present. The posterior clears the floor, so the veto does not engage and no rescue is needed. Consistent with \cref{sec_l3_decomp}, L3 passes the decision through unchanged.

\noindent\textbf{Stage 3: decision.} Compromise, matching SME adjudication. The human reviewer had read the established funding business and long-lived domain as reassuring; the stage weighed those against the absent baseline history and the simultaneous language and country shift, which is the conjunction its rules encode.

\section{Agent Configuration, Prompt Sketch, and Output Schema}
\label{app_prompt}

\paragraph{Agent specification.} \Cref{tab_agent_config} gives the full configuration of the agent described in \cref{sec_llm_agent}.

\begin{table}[h]
\centering
\caption{LLM agent specification.}
\label{tab_agent_config}
\small
\begin{tabularx}{\linewidth}{lX}
\toprule
\textbf{Component} & \textbf{Setting} \\
\midrule
Backbone & Gemini 2.5 Flash (hosted serving platform) \\
Tool interface & native function calling; tools discovered at runtime via MCP \\
Tools registered & 31 tools produce the signal vector \\
Reasoning loop & multi-turn tool calling; terminates on a final answer with no tool call \\
Step cap & 25 \\
Decoding & \texttt{max\ub output\ub tokens} $=65536$; temperature at model default \\
Determinism & identical tool calls memoized within a run \\
Structured output & one JSON object: \texttt{final\ub decision} $\in\{$Compromise, Not Compromise$\}$, \texttt{confidence}, free-text \texttt{reasoning}, 7 \texttt{override\ub checks} booleans, 5 \texttt{evidence} booleans, 10 per-component \texttt{checks} each with \texttt{verdict} $\in\{$suspicious, normal, inconclusive$\}$ \\
Parsing & tolerates fenced or bare JSON, with a re-trying fallback \\
\bottomrule
\end{tabularx}
\end{table}

\paragraph{The prompt is not reproduced verbatim.} We give a structural sketch sufficient to re-implement the \emph{method}. We flag this as a genuine reproducibility limitation rather than presenting the sketch as equivalent to the prompt.

\paragraph{Structure of the system prompt.} The prompt is assembled in seven blocks, in order:
\begin{enumerate}\itemsep0pt
\item \textbf{Task framing and persona}: the reviewer role, and the ad under review.
\item \textbf{The central distinction}: compromise (a legitimate account taken over) versus abuse (a bad-faith account from the outset) versus a legitimate behavioral shift, with an explicit \emph{"default to Not Compromise"} prior instructing the model to require clear, strong evidence before predicting compromise.
\item \textbf{Adversary monetization taxonomy}: an enumeration of the ways an attacker converts stolen ad credit into money, each with the observable signature that distinguishes it from the same behavior performed legitimately. \emph{Content withheld.}
\item \textbf{Per-component analysis guides}: for each of the nine review components (content, creator, page, target countries, destination URL, pixel, payment, business connections, ad-account connections), the questions the reviewer must answer, grouped into recency analysis (is this artefact new to this business?) and connection analysis (does this artefact's wider association graph fit this business?).
\item \textbf{Initial assessment}: the upstream classifier's score and features, injected with an explicit instruction that it is a starting point only, must not be the sole basis for a decision, and must be checked against independent tool evidence.
\emph{Conditions and thresholds withheld.}
\item \textbf{Output contract}: an instruction that each check verdict be evaluated independently on that check's own evidence rather than back-filled from the final decision, followed by the JSON shown in figure \ref{lst_schema}.
\end{enumerate}
A managed-tier annotation is appended when the business has one, instructing the model to require stronger evidence because false positives might be more relevant in certain cases.

\paragraph{Output schema.} The agent emits a single JSON object. \Cref{lst_schema} sketches its shape. The per-component checks are abstracted as \texttt{sme\ub check\ub i}, the shift evidence is shown as a template over the dimension it applies to, and only the override checks already named elsewhere in this paper appear.

\begin{figure}[t]
\centering
{\scriptsize
\begin{verbatim}
{
  "final_decision":
      "Compromise" | "Not Compromise",
  "reasoning": "<free-text rationale>",
  "confidence": "high" | "medium" | "low",

  "override_checks": {
    "is_abusive_not_compromised": <bool>,
    "funding_business_is_established":
        <bool>,
    "has_sufficient_baseline_history":
        <bool>,
    ...
  },

  "evidence": {
    "<dimension>_shift_detected": <bool>,
    "creator_is_original_to_business":
        <bool>,
    ...
  },

  "checks": {
    "sme_check_1": {
      "verdict": "suspicious" | "normal"
                 | "inconclusive",
      "<check_specific_signal>": <bool>
    },
    "sme_check_2": { ... },
    ...
  }
}
\end{verbatim}}
\caption{Sketch of the agent's JSON response. One \texttt{checks} object is emitted per component of the SME review checklist, each carrying a verdict and its own boolean sub-signals; \texttt{<dimension>} ranges over the observable properties of an ad whose shift the reviewer tracks.}
\label{lst_schema}
\end{figure}

The boolean sub-signals are instructed to reflect the model's semantic judgement after reading the tool outputs, not a restatement of any single tool's field. Parsing accepts a fenced or bare JSON object with a retry fallback, and falls back to text-based extraction of the final decision if no JSON can be recovered; every field has a default, so a partial response degrades rather than fails. The signal vector consumed by the decision stage is this JSON, the per-component verdicts and their sub-signals, the override and evidence booleans, the confidence level, and the count of suspicious verdicts, concatenated with the deterministic numeric signals returned by the tools.

\section{Operational Distributions}
\label{app_cost}
\Cref{tab_cost} gives the measured per-review cost summarized in \cref{sec_cost}. Token consumption is modest in absolute terms, which is what makes the design affordable despite the multi-turn loop. The token and latency distributions below let a reader derive it for their own execution, which is the portable form of the number.

\begin{table}[h]
\centering
\caption{Measured operational characteristics over 330 samples. Agent latency is end-to-end; stage latency is the neuro-symbolic decision stage. Reasoning steps are not logged separately; tool calls are the logged proxy and are bounded by the 25-step cap.}
\label{tab_cost}
\small
\setlength{\tabcolsep}{4pt}
\begin{tabular}{@{}lrrrr@{}}
\toprule
\textbf{Quantity} & \textbf{Mean} & \textbf{p50} & \textbf{p90} & \textbf{p99} \\
\midrule
Tool calls & 3.6 & 1 & 10 & 16 \\
Agent latency (s) & 175.5 & 172.1 & 242.5 & 330.0 \\
Logic Stage latency (ms) & 20.7 & 15.6 & 17.4 & 25.2 \\
Est.\ tokens & 742 & 770 & 1029 & 1373 \\
\bottomrule
\end{tabular}
\end{table}

\section{Over-sampled General Population Results}
\label{app_cv}
\Cref{tab:cv-simulated} gives the 5-fold cross-validation results discussed in \cref{sec_exp_cross_validate}.

\begin{table}[h]
  \centering
  \caption{5-fold CV on the over-sampled general population (975 cases, 684 positive, $\sim$70\%; compromise over-sampled). The always-positive row predicts compromise for every case. Tree baselines use the same signal vector.}
  \label{tab:cv-simulated}
\small
\setlength{\tabcolsep}{4pt}
  \begin{tabular}{@{}lcccc@{}}
  \toprule
  \textbf{Method} & \textbf{Prec} & \textbf{Rec} & \textbf{F1} & \textbf{MCC} \\
  \midrule
  Always-positive & 0.702 & 1.000 & 0.825 & 0.000 \\
  LLM Agent alone & 0.809 & 0.953 & 0.875 & 0.513 \\
  Decision Tree & 0.921 & 0.920 & 0.920 & 0.733 \\
  ILP+NB (L1+L2) & 0.916 & 0.939 & 0.927 & 0.749 \\
  FOIL-IE (L1) & 0.926 & 0.931 & 0.929 & 0.759 \\
  Gradient Boosting & 0.921 & 0.966 & 0.943 & 0.800 \\
  Neuro-symbolic (ours) & \textbf{0.951} & 0.931 & 0.941 & 0.807 \\
  Random Forest & 0.932 & \textbf{0.966} & \textbf{0.949} & \textbf{0.823} \\
  \bottomrule
  \end{tabular}
\end{table}

\section{Full Hard Cases Population}
\label{app_full397}
\Cref{tab_full397} repeats the comparison on all 397 SME-adjudicated accounts (60 positives), without the human baseline. The ordering is compressed relative to \cref{tab:ablation-train}, the full stage, FOIL-IE alone, and random forest are separated by less than 0.005 MCC, reinforcing that these methods are statistically indistinguishable at this label volume.

\begin{table}[t]
\centering
\caption{All 397 SME-adjudicated accounts (60 positives, $15.1\%$). Same folds, gating, and per-fold threshold tuning as \cref{tab:ablation-train}. SD is across the 5 CV test folds.}
\label{tab_full397}
\small
\setlength{\tabcolsep}{3pt}
\begin{tabular}{@{}lcccc@{}}
\toprule
\textbf{Method} & \textbf{Prec} & \textbf{Rec} & \textbf{F1} & \textbf{MCC (SD)} \\
\midrule
Agent alone & 0.248 & 0.917 & 0.390 & 0.304 (0.087) \\
L2: NB only & 0.414 & 0.483 & 0.446 & 0.340 (0.143) \\
Gradient Boosting & 0.457 & 0.350 & 0.396 & 0.309 (0.133) \\
L1+L2: ILP+NB & 0.350 & 0.683 & 0.463 & 0.360 (0.098) \\
Random Forest & 0.437 & 0.517 & 0.473 & 0.372 (0.082) \\
L1: FOIL-IE & 0.384 & 0.633 & 0.478 & 0.374 (0.093) \\
Full (ours) & 0.400 & 0.600 & 0.480 & 0.376 (0.121) \\
\bottomrule
\end{tabular}
\end{table}

\section{Contradiction Types and Resolution}
\label{app_contra}
\begin{table*}[t]
\centering
\caption{Contradiction types and how the resolution layer acts on them.}
\label{tab:contradiction-types}
\small
\begin{tabularx}{\linewidth}{l>{\raggedright\arraybackslash}X}
\toprule
\textbf{Type} & \textbf{Resolution} \\
\midrule
Compromise leaning & Rescue a negative prediction if the NB posterior stays above a secondary threshold; override an NB veto when L1 is positive. \\
False-negative leaning & Rescue when neither L1 nor L2 fires but hard indicators (compromised domain/page/pixel) persist and the posterior is above a rescue threshold. \\
False-positive leaning & Veto a positive when conflicting evidence is present, the posterior is uncertain, and no compromise-leaning evidence counterbalances it. \\
Direct rule contradiction & L1 priority: high-confidence confirmations win, else rejections; the flag is also an NB feature. \\
Agent-rejection contradiction & L1 rejection overrides the agent's positive verdict (rules are grounded in deterministic signals); also an NB feature. \\
\bottomrule
\end{tabularx}
\end{table*}

\end{document}